\documentclass[runningheads,a4paper]{llncs}

\usepackage{amssymb}
\usepackage{graphicx}
\usepackage[ruled,vlined,linesnumbered]{algorithm2e}
\usepackage{float}
\usepackage{color}
\usepackage{soul}
\usepackage{amsmath}
\usepackage{caption}

\usepackage{url}

\urldef{\mailsa}\path|{msarker, acolman, jhan}@swin.edu.au}|
\urldef{\mailsb}\path|{akabir@csu.edu.au}|
\urldef{\mailsc}\path|erika.siebert-cole, peter.strasser, lncs}@springer.com|    
\newcommand{\keywords}[1]{\par\addvspace\baselineskip
\noindent\keywordname\enspace\ignorespaces#1}

\begin{document}

\mainmatter  

\title{An Improved Naive Bayes Classifier-based Noise Detection Technique for Classifying User Phone Call Behavior}

\titlerunning{An Improved Naive Bayes Classifier-based Noise Detection Technique}

%
%

\author{Iqbal H. Sarker$^1$%
\thanks{Springer International Publishing, Yee Ling Boo et al. (Eds.). This paper appeared at the Fifteenth Australasian Data Mining Conference (AusDM 2017), Melbourne, Australia.}%
\and Muhammad Ashad Kabir$^2$ \and Alan Colman$^1$ \and Jun Han$^1$}
\authorrunning{Iqbal H. Sarker et al.}

\institute{\textsuperscript 1 Department of Computer Science \& Software Engineering,\\
	School of Software and Electrical Engineering,\\
	Swinburne University of Technology,
	Melbourne, Australia\\
	\mailsa\\
	\textsuperscript 2 School of Computing and Mathematics,\\
	Charles Sturt University, NSW, Australia\\    
	\mailsb\\
}

%
%

\toctitle{Lecture Notes in Computer Science}
\tocauthor{Authors' Instructions}
\maketitle

\begin{abstract}
The presence of noisy instances in mobile phone data is a fundamental issue for classifying user phone call behavior (i.e., accept, reject, missed and outgoing), with many potential negative consequences. The classification accuracy may decrease and the complexity of the classifiers may increase due to the number of redundant training samples. To detect such noisy instances from a training dataset, researchers use naive Bayes classifier (NBC) as it identifies misclassified instances by taking into account independence assumption and conditional probabilities of the attributes. However, some of these misclassified instances might indicate \textit{usages behavioral patterns} of individual mobile phone users. Existing naive Bayes classifier based noise detection techniques have not considered this issue and, thus, are lacking in \textit{classification accuracy}. In this paper, we propose an \textit{improved noise detection technique} based on naive Bayes classifier for effectively classifying users' phone call behaviors. In order to improve the classification accuracy, we effectively identify noisy instances from the training dataset by analyzing the behavioral patterns of individuals. We dynamically determine a \textit{noise threshold} according to individual's unique behavioral patterns by using both the naive Bayes classifier and Laplace estimator. We use this noise threshold to identify noisy instances. To measure the effectiveness of our technique in classifying user phone call behavior, we employ the most popular classification algorithm (e.g., decision tree). Experimental results on the real phone call log dataset show that our proposed technique more accurately identifies the noisy instances from the training datasets  that leads to better classification accuracy.

\keywords{Mobile Data Mining, Noise Analysis, Naive Bayes Classifier, Decision Tree, Classification, Laplace Estimator, Predictive Analytics, Machine Learning, User Behavior Modeling.}
\end{abstract}

\section{Introduction}
Now a days, mobile phones have become part of our daily life. The number of mobile cellular subscriptions is almost equal to the number of people on the planet \cite{pejovic2014interruptme} and the phones are, for most of the day, with their owners as they go through their daily routines \cite{pejovic2014interruptme}. People use mobile phones for various activities such as voice communication, Internet browsing, app using, e-mail, online social network, instant messaging, etc. \cite{pejovic2014interruptme}. In recent years, researchers use various types of mobile phone data such as phone call log \cite{ozer2016predicting}, app usages log \cite{srinivasan2014mobileminer}, mobile phone notifications history \cite{mehrotra2016prefminer}, web log \cite{halvey2005time}, context log \cite{zhu2014mining} for different personalized applications. For instance, phone call log is used to predict users' behavior in order to build an automated call firewall or call reminder system~\cite{phithakkitnukoon2011behavior}.

In data mining area, classification is a function that describes and distinguishes data classes or concepts \cite{farid2014hybrid}. The goal of classification is to accurately classify the class labels of instances whose attribute values are known, but class values are unknown. Accurately classifying user phone call behavior from log data using machine learning techniques (e.g., decision tree) is challenging as it requires a data set free from outliers or noise \cite{daza2007algorithm}. However, real-world datasets may contain noise, which is anything that obscures the relationship between the features of an instance and it's behavior class \cite{frenay2014classification}. Such noisy instances may reduce the classification accuracy, and increase the complexity of the classification process. It is also evident that decision trees are badly impacted by noise \cite{frenay2014classification}. Hence, we summarize the effects of noisy instances for classifying user phone call behavior as follows:
 
\begin{itemize}
	\item Create unnecessary classification rules that are not interesting to the users and make the rule-set larger.
	\item The complexity of the classifiers and the number of necessary training samples may increase.
	\item The presence of noisy training instances is more likely to cause  over-fitting for the decision tree classifier and thus decrease it's accuracy.
\end{itemize}

According to \cite{zhu2004class}, the performance of the classifier depends on two significant factors: (1) the quality of the training data, and (2) the competence of learning algorithm. Therefore, identification and elimination of the noisy instances from a training dataset are required to ensure the quality of the training data before applying learning technique in order to achieve better classification accuracy. 

NBC is the most popular technique to detect noisy instances from a training dataset, as it is attributed to the independence assumption and the use of conditional probabilities \cite{farid2014hybrid} \cite{chen2009feature}. Farid et al. \cite{farid2014hybrid} have proposed a naive Bayes classifier based noise detection technique for multi-class classification tasks. This technique finds the noisy instances from a training dataset using a naive Bayes classifier and removes these instances from the training set before constructing a decision tree learning for making decisions. In their approach, they identify all the misclassified instances from the training dataset using NBC and consider these instances as noise.  However, some of these misclassified instances  might represent true\textit{ behavioral patterns} of individuals. Therefore, such a strong assumption regarding noisy instances  more likely to decrease the \textit{classification accuracy} of mining phone call behavior.

In this paper, we address the above mentioned issue for identifying noisy instances and propose an \textit{improved noise detection technique} based on the naive Bayes classifier for effectively classifying mobile users' phone call behaviors. In our approach, we first calculate the conditional probability for all the instances using naive Bayes classifier and Laplace-estimator. After that we dynamically determine a \textit{noise threshold} according to individual's unique behavioral patterns. Finally, the (misclassified) instances that can't satisfy this threshold are selected as noise. As individual's phone call behavioral patterns are not identical in the real life, this threshold for identifying noisy instances changes dynamically according to the behavior of individuals. To measure the effectiveness of our technique for classifying user phone call behavior, we employ a prominent classification algorithm - decision tree. Our approach aims to improve the existing naive Bayes classifier based noise detection technique \cite{farid2014hybrid} for classifying phone call behavior of individuals. \\

The contributions are summarized as follows:
\begin{itemize}
	\item We determine a \textit{noise threshold} dynamically according to individual's unique behavioral patterns.
	
	\item We propose an \textit{improved noise detection technique} based on naive Bayes classifier for effectively classifying mobile users' phone call behaviors. 
	
	\item Our experiments on real mobile phone datasets show that this technique is more effective than existing technique for classifying user phone call behavior.   
\end{itemize}

The rest of the paper is organized as follows. We review the naive Bayes classifier and Laplacian estimator in Section \ref{Naive Bayes classifier} and Section \ref{Laplacian Estimation} respectively. We present our approach in section \ref{Our Noise Detection Technique}. We report the experimental results in Section \ref{Experiments}. Finally, Section \ref{Conclusion and Future Work} concludes this the paper and highlights the future work.

\section{Naive Bayes Classifier}
\label{Naive Bayes classifier}
A naive Bayes classifier (NBC) is a simple probabilistic based method, which can predict the class membership probabilities \cite{chen2009feature} \cite{han2011data}. It has two main advantages: (a) easy to use, and (b) only one scan of the training data is required for probability generation. A naive Bayes classifier can easily handle missing attribute values by simply omitting the corresponding probabilities for those attributes when calculating the likelihood of membership for each class. It also requires the class conditional independence, i.e., the effect of an attribute on a given class is independent of those of other attributes.

Let D be a training set of data instances and their associated class labels. Each instance is represented by an n-dimensional attribute vector, $X = \{x_1, x_2,..., x_n\}$, depicting $n$ measurements made on the instance from $n$ attributes, respectively, $\{A_1, A_2,..., A_n\}$. Suppose that there are $m$ classes, $\{C_1, C_2,..., C_m\}$. For a test instance, $X$, the classifier will predict that $X$ belongs to the class with the highest conditional probability, conditioned on $X$. That is, the naive Bayes classifier predicts that the instance $X$ belongs to the class $C_i$, if and only if - \\

$P(C_i|X) > P(C_j|X)$ for $1 \leq j \leq m, j \neq i$ \\

The class $C_i$ for which $P(C_i|X)$ is maximized is called the Maximum Posteriori Hypothesis.

\begin{equation}
	P(C_i|X) =\frac{P(X|C_i)P(C_i)}{P(X)}
\end{equation}

In Bayes theorem shown in Equation (1), as $P(X)$ is a constant for all classes, only $P(X|C_i)P(C_i)$ needs to be maximized. If the class prior probabilities are not known, then it is commonly assumed that the classes are likely equal, that is, $P(C_1) = P(C_2) = ... = P(C_m)$, and therefore we would maximize $P(X|C_i)$. Otherwise, we maximize $P(X|C_i)P(C_i)$. The class prior probabilities are calculated by $P(C_i) = |C_{i,D}| / |D|$, where $|C_{i,D}|$ is the number of training instances of class $C_i$ in $D$. To compute $P(X|C_i)$ in a dataset with many attributes is extremely computationally expensive. Thus, the naive assumption of class-conditional independence is made in order to reduce computation in evaluating $P(X|C_i)$. This presumes that  the attributes' values are conditionally independent of one another, given the class label of the instance, i.e., there are no dependence relationships among attributes. Thus, Equation (2) and (3) are used to produce $P(X|C_i)$.

\begin{equation}
P(X|C_i) = \prod_{k=1}^{n} P(x_k|C_i)
\end{equation}
\begin{equation}
P(X|C_i) = P(x_1|C_i) \times P(x_2|C_i) \times ... \times P(x_n|C_i)
\end{equation}

In Equation (2), $x_k$ refers to the value of attribute $A_k$ for instance $X$. Therefore, these probabilities $P(x_1|C_i), P(x_2|C_i),..., P(x_n|C_i)$ can be easily estimated from the training instances. If the attribute value, $A_k$, is categorical, then $P(x_k|C_i)$ is the number of instances in the class $C_i \in D$ with the value $x_k$ for $A_k$, divided by $|C_{i,D}|$, i.e., the number of instances belonging to the class $C_i \in D$.

To predict the class label of instance $X, P(X|C_i)P(C_i)$ is evaluated for each class $C_i \in D$. The naive Bayes classifier predicts that the class label of instance $X$ is the class $C_i$, if and only if - \\

$P(X|C_i)P(C_i) > P(X|C_j)P(C_j)$ for $1 \leq j \leq m$ and $j \neq i$

In other words, the predicted class label is the class $C_i$ for which $P(X|C_i)P(C_i)$ is the maximum.

\section{Laplacian Estimation}
\label{Laplacian Estimation}
As in naive Bayes classifier, we calculate $P(X|C_i)$ as the product of the probabilities $P(x_1|C_i) \times P(x_2|C_i) \times ... \times P(x_n|C_i)$, based on the independence assumption and class conditional probabilities, we will end up with a probability value of zero for some $P(x|C_i)$ if attribute value $x$ is never observed in the training data for class $C_i$. Therefore, Equation (3) becomes zeros for such attribute value regardless the values of other attributes. Thus, naive Bayes classifier cannot predict the class of such test instance. Laplace estimate \cite{cestnik1990estimating} is usually employed to scale up the values by smoothing factor. In Laplace-estimate, the class probability is defined as:

\begin{equation}
	P(C = c_i) = \frac{n_c + k}{N + n \times k}
\end{equation}

where $n_c$ is the number of instances satisfying $C = c_i$, $N$ is the number of training instances, $n$ is the number of classes and $k = 1$.

Let's consider a phone call behavior example, for the behavior class `reject' in the training data containing 1000 instances, we have 0 instance with $relationship = unknown$, 990 instances with $relationship = friend$, and 10 instances with $relationship = mother$. The probabilities of these contexts are 0, 0.990 (from 990/1000), and 0.010 (from 10/1000), respectively. On the other hand, according to equation (4), the probabilities of these contexts would be as follows:\\

$\frac{1}{1003} = 0.001$,
$\frac{991}{1003} = 0.988$,
$\frac{11}{1003} = 0.011$

\bigskip In this way, we obtain the above non-zero probabilities (rounded up to three decimal places) respectively using Laplacian-estimation. The ``new'' probability estimates are close to their ``previous'' counterparts, and these values can be used for further processing.

\section{Noise Detection Technique}
\label{Our Noise Detection Technique}
\begin{figure}
	\centering
	\includegraphics[height=5.5cm,keepaspectratio]{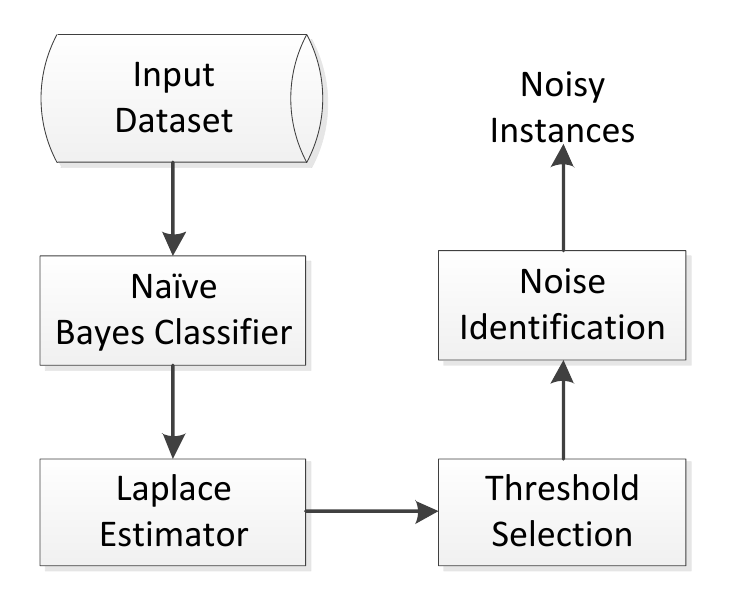}
	\caption{A block diagram of noise detection technique}
	\label{fig:overview}
\end{figure}

In this section, we discuss our noise detection technique in order to effectively classify user phone call behavior. Figure \ref{fig:overview} shows the block diagram of our noise detection technique. 

In order to detect noise, we use naive Bayes classifier (NBC) \cite{han2011data} as the basis for noise identification. Using NBC, we first calculate the conditional probability for each attribute by scanning the training data. Table \ref{sample-datasets} shows an example of the mobile phone dataset. Each instance contains four attribute values (e.g., time, location, situation, and  relationship between caller and callee) and corresponding phone call behavior. Table \ref{prior-probability} and Table \ref{conditional-probability} report the prior probabilities for each behavior class and conditional probabilities for each attribute value, respectively for this dataset. Using these probabilities, we calculate the conditional probability for each instance. As NBC was implemented under the independence assumption, it estimates zero probabilities if the conditional probability for a single attribute is zero. In such cases, we use Laplace-estimator \cite{cestnik1990estimating} to estimate the conditional probability of any of the attribute value. 

\begin{table*}[htbp!]
	\centering
	\caption{Sample mobile phone dataset}
	\label{sample-datasets}
	\begin{tabular}{|c|c|c|c|c|} 
		\hline
		\bf Day[Time-Segment] & \bf Location & \bf Situation & \bf Relationship & \bf User Behavior \\  
		\hline
		Fri[S1] & Office & Meeting & Friend & Reject \\ 
		Fri[S1] & Office & Meeting & Colleague & Reject \\ 
		Fri[S1] & Office & Meeting & Boss & Accept \\ 
		Fri[S1] & Office & Meeting & Friend & Reject \\
		Fri[S2] & Home & Dinner & Friend & Accept \\ 
		Wed[S1] & Office & Seminar & Unknown & Reject \\
		Wed[S1] & Office & Seminar & Colleague & Reject \\
		Wed[S1] & Office & Seminar & Mother & Accept \\
		Wed[S2] & Home & Dinner & Unknown & Accept \\ 
		\hline
	\end{tabular}
\end{table*}

\begin{table*}[htbp!]
	\centering
	\caption{Prior probabilities for each behavior class generated using the mobile phone dataset}
	\label{prior-probability}
	\begin{tabular}{|c|c|} 
		\hline
		\bf Probability & \bf Value \\  
		\hline
		P(behavior = Reject) & 5/9  \\ 
		P(behavior = Accept) & 4/9  \\ 
		\hline
	\end{tabular}
\end{table*}

\begin{table*}[htbp!]
	\centering
	\caption{Conditional probabilities for each attribute value calculated using the mobile phone dataset}
	\label{conditional-probability}
	\begin{tabular}{|c|c|} 
		\hline
		\bf Probability & \bf Value \\  
		\hline
		$P(DayTime = Fri[S1] | behavior = Reject)$ & 3/5  \\ 
		$P(DayTime = Fri[S1] | behavior = Accept)$ & 1/4  \\ 
		$P(DayTime = Fri[S2] | behavior = Reject)$ & 0/5  \\ 
		$P(DayTime = Fri[S2] | behavior = Accept)$ & 1/4  \\ 
		$P(DayTime = Wed[S1] | behavior = Reject)$ & 2/5  \\ 
		$P(DayTime = Wed[S1] | behavior = Accept)$ & 1/4  \\ 
		$P(DayTime = Wed[S2] | behavior = Reject)$ & 0/5  \\ 
		$P(DayTime = Wed[S2] | behavior = Accept)$ & 1/4  \\ 
		$P(Location = Office | behavior = Reject)$ & 5/5  \\ 
		$P(Location = Office | behavior = Accept)$ & 2/4  \\ 
		$P(Location = Home | behavior = Reject)$ & 0/5  \\ 
		$P(Location = Home | behavior = Accept)$ & 2/4  \\ 
		$P(Situation = Meeting | behavior = Reject)$ & 3/5  \\ 
		$P(Situation = Meeting | behavior = Accept)$ & 1/4  \\ 
		$P(Situation = Seminar | behavior = Reject)$ & 2/5  \\ 
		$P(Situation = Seminar | behavior = Accept)$ & 1/4  \\ 
		$P(Situation = Dinner | behavior = Reject)$ & 0/5  \\ 
		$P(Situation = Dinner | behavior = Accept)$ & 2/4  \\ 
		$P(Relationship = Friend | behavior = Reject)$ & 2/5  \\ 
		$P(Relationship = Friend  | behavior = Accept)$ & 1/4  \\ 
		$P(Relationship = Colleague | behavior = Reject)$ & 2/5  \\ 
		$P(Relationship = Colleague  | behavior = Accept)$ & 0/4  \\ 
		$P(Relationship = Boss | behavior = Reject)$ & 0/5  \\ 
		$P(Relationship = Boss  | behavior = Accept)$ & 1/4  \\ 
		$P(Relationship = Mother | behavior = Reject)$ & 0/5  \\ 
		$P(Relationship = Mother  | behavior = Accept)$ & 1/4  \\ 
		$P(Relationship = Unknown | behavior = Reject)$ & 1/5  \\ 
		$P(Relationship = Unknown  | behavior = Accept)$ & 1/4  \\ 
		\hline
	\end{tabular}
\end{table*}

\begin{figure}[htbp!]
	\centering
	\includegraphics[width=.7\linewidth,keepaspectratio]{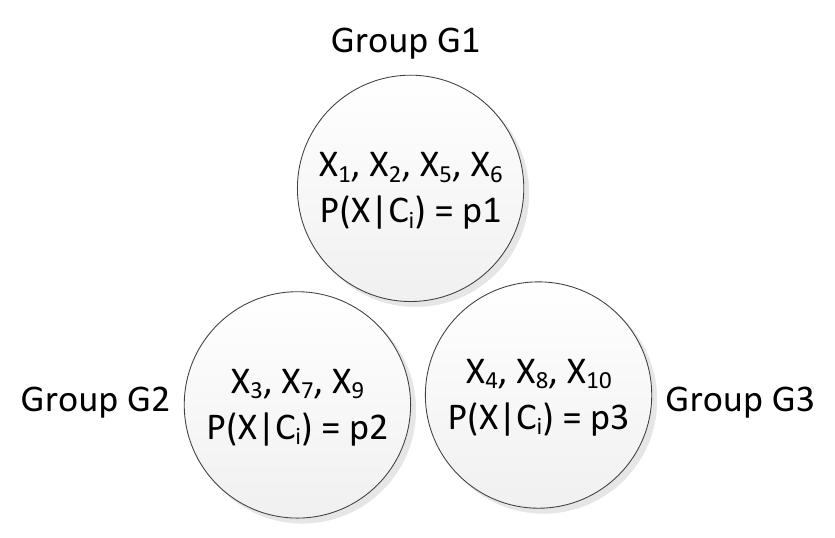}
	\caption{An example of instances-group based on probability}
	\label{fig:group}
\end{figure}

Once we have calculated conditional probability for each instance, we differentiate between the purely classified instances and misclassified instances using these values. ``Purely classified'' instances are those for which the predicted class and the original class is same. If different class found then these are ``misclassified'' instances. After that, we generate the instances groups by taking into account all the distinct probabilities as separate group values. Figure \ref{fig:group} shows an example of instances groups $G1, G2, G3$ for the instances ${X_1,X_2,...,X_{10}}$ where $G1$ consists of 5 instances with probability $p1$, $G2$ consists of 3 instances with probability $p2$ and finally $G3$ consists of 3 instances with probability $p3$. We then identify the group among the purely classified instances for which the probability is minimum. This minimum probability is considered as ``noise-threshold''. Finally, the instances in misclassified list, for those probabilities are less than  the noise threshold, are identified as noise.

\begin{algorithm}[htbp!]
	\caption{Noise Detection}
	\label{alg:noise-detection}
	\SetKwInOut{Data}{Data}
	\SetAlgoLined
	\Data{Training dataset: $D = {X_1,X_2,...,X_n}$ // Training dataset, $D$, which contains
	a set of training instances and their associated class labels.}
	\KwResult{noise list: $noise_{list}$}
	
	\BlankLine
	
	\ForEach{class, $C_i \in D$}
	{  
		Find the prior probabilities, $P(C_i)$.	
	}
	\ForEach{attribute value, $A_{i} \in D$}
	{  
		Find the class conditional probabilities, $P(A_{i}|C_i)$.	
	}
	\ForEach{training instance, $X_i \in D$}
	{  
		Find the conditional probability, $P(X_i|C_i)$ \\
		
		\If{$P(X_i|C_i)$ $==$ 0}
		{
			//use Laplacian Estimator\\
			recalculate the conditional probability, $P(X_i|C_i)$ using Laplacian Estimator
		}
		\If{$X_i$ is misclassified}
		{
			 $misClass_{list} \leftarrow X_i$ \\
			 $misPro_{list} \leftarrow P(X_i|C_i)$	// store the probabilities for all misclassified instances.  
		}
		\Else
		{
	 		$pureClass_{list} \leftarrow X_i$ \\
	 		$purePro_{list} \leftarrow P(X_i|C_i)$	// store the probabilities for all purely classified instances.   
		}
	}

	$T_{noise} $=$ findMIN(purePro_{list})$ // use as noise threshold	
	
	\ForEach{instance, $x_i \in misClass_{list}$}
	{  
		Find the conditional probability, $P(X_i|C_i)$ from $misPro_{list}$
		
		\If{$P(X_i|C_i)$ $<$ $T_{noise}$} 
		{
			$noise_{list} \leftarrow X_i$ // store instances as noise.
		}
	}
	return $noise_{list}$
	
\end{algorithm}

The process for identifying noise is set out in Algorithm \ref{alg:noise-detection}. Input data includes training dataset: $D = {X_1,X_2,...,X_n}$, which contains a set of training instances and their associated class labels and output data is the list of noisy instances. For each class, we calculate the prior probabilities $P(C_i)$ (line 2). After that for each attribute value, we calculate the class conditional probabilities $P(A_{i}|C_i)$ (line 5). For each training instance, we calculate the conditional probabilities $P(X_i|C_i)$ (line 8). We then check whether it is non-zero. If we get zero probabilities, we then recalculate the conditional probabilities $P(X_i|C_i)$ using Laplacian Estimator (line 11). Based on these probability values, we then check whether the instances are misclassified or purely classified and store all misclassified instances $misClass_{list}$ (line 14) with corresponding probabilities in $misPro_{list}$ (line 15). Similarly, we also store all purely classified instances $pureClass_{list}$ (line 18) with corresponding probabilities in $purePro_{list}$ (line 19). We then identify the minimum probability from $purePro_{list}$ as noise threshold (line 22). As we aim to identify the noise list we check the conditional probabilities $(X_i|C_i)$ in $misPro_{list}$ for all instances. If any instance fails to satisfy this threshold then we store that instance as noise and store into $noise_{list}$ (line 26). Finally this algorithm returns a set of noisy instances $noise_{list}$ (line 29) for a particular dataset.

Rather than arbitrarily determine the threshold, our algorithm dynamically identifies the noise threshold according to individual's behavioral patterns and identify noisy instances based on this threshold. As individual's phone call behavioral patterns are not identical in the real life this noise-threshold for identifying noisy instances changes dynamically according to individual's unique behavioral patterns.  

\section{Experiments}
\label{Experiments}
In this section, we describe our experimental setup and the phone log datasets used in experiment. We also present an experimental evaluation comparing our proposed noise detection technique and the existing naive Bayes classifier based noise detection technique \cite{farid2014hybrid} for classifying user phone call behavior.

\subsection{Experimental Setup}
We have implemented our noise detection technique (Algorithm \ref{alg:noise-detection}) and existing naive Bayes classifier based technique \cite{farid2014hybrid} in Java programming language and executed them on a Windows PC with an Intel Core I5 CPU (3.20GHz) and 8GB memory. In order to measure the classification accuracy, we first eliminate the noisy instances identified by noise identification technique from the training dataset, and then apply the decision-tree classifier \cite{quinlan1993} on the noise-free dataset. The reason for choosing the decision tree as a classifier is that decision tree is the most popular classification algorithm in data mining \cite{wu2008top} \cite{wu2016decision}. The code for the basic versions of the decision tree classifier is adopted from Weka, which is an open source data mining software \cite{hall2009weka}.

\subsection{Dataset}
We have conducted experiments on phone log datasets of five individual mobile phone users (randomly selected from Massachusetts Institute of Technology (MIT) Reality Mining dataset~\cite{eagle2006infering}). We extract 7-tuple information of the call record for each phone user from the datasets: {date of call, time of call, call-type, call duration, location, relationship, call ID}. These datasets contain three types of phone call behavior, e.g., incoming, missed and outgoing. As can be seen, the user's behavior in accepting and rejecting calls are not directly distinguishable in incoming calls in the dataset. As such, we derive accept and reject calls by using the call duration. If the call duration is greater than 0 then the call has been accepted; if it is equal to 0 then the call has been rejected \cite{sarker2016behavior}. We also pre-process the temporal data in mobile phone log as it is continuous and numeric. For this, we use BOTS technique \cite{sarker2016behavior} for producing behavior-oriented time segments. Table \ref{Datasets descriptions} describes each dataset of the individual mobile phone user.

\begin{table}[htbp!]
	\centering
	\caption{Datasets descriptions}
	\label{Datasets descriptions}
	\begin{tabular}{|c|c|c|c|} 
		\hline
		\bf Dataset & \bf Contexts & \bf Instances & \bf Behavior Classes \\  
		\hline
		User 04 & temporal, location, relationship & 5119 & accept, reject, missed, outgoing \\ 
		\hline
		User 23 & temporal, location, relationship & 1229 & accept, reject, missed, outgoing \\ 
		\hline
		User 26 & temporal, location, relationship & 3255 & accept, reject, missed, outgoing \\ 
		\hline
		User 33 & temporal, location, relationship & 635 & accept, reject, missed, outgoing \\ 
		\hline
		User 51 & temporal, location, relationship & 2096 & accept, reject, missed, outgoing \\ 
		\hline
	\end{tabular}
\end{table}

\subsection{Evaluation Metric}
In order to measure the classification accuracy, we compare the classified response with the actual response (i.e., the ground truth) and compute the accuracy in terms of:

\begin{itemize}
	\item Precision: ratio between the number of phone call behaviors that are correctly classified and the total number of behaviors that are classified (both correctly and incorrectly). If TP and FP denote true positives and false positives then the formal definition of precision is:
	
	\begin{equation}
	Precision = \frac{TP}{TP + FP}
	\end{equation}
	
	\item Recall: ratio between the number of phone call behaviors that are correctly classified and the total number of behaviors that are relevant. If TP and FN denote true positives and false negatives then the formal definition of recall is:
	
	\begin{equation}
	Recall = \frac{TP}{TP + FN}
	\end{equation}
	
	\item F-measure: a measure that combines precision and recall is the harmonic mean of precision and recall. The formal definition of F-measure is:
	
	\begin{equation}
	Fmeasure = 2 * \frac{Precision * Recall}{Precision + Recall}
	\end{equation}
	
\end{itemize}

\subsection{Evaluation Results}
To evaluate our approach, we employ the 10-fold cross validation on each dataset. In $k$ fold cross-validation, the initial data are randomly partitioned into $k$ mutually exclusive subsets or ``folds'', $d_1,d_2,...,d_k$, each of which has an approximately equal size. Training and testing are performed $k$ times. In iteration $i$, the partition $d_i$ is reserved as the test set, and the remaining partitions are collectively used to train the classifier. Therefore, the 10-fold cross validation breaks data into 10 sets of size N/10. It trains the classifier on 9 sets and tests it using the remaining one set. This repeats 10 times and we take a mean accuracy rate. For classification, the accuracy estimate is the total number of correct classifications from the k-iterations, divided by the total number of instances in the initial dataset. To show the effectiveness of our technique, we compare the accuracy of both the existing naive Bayes classifier based noise detection approach (NBC) \cite{farid2014hybrid} and our proposed dynamic threshold based approach, in terms of precision, recall and f-measure.
  
\begin{table}[htbp!]
	\centering
	\caption{The accuracies of existing naive Bayes classifier based approach (NBC)}
	\label{existing-naive}
	\begin{tabular}{|c|c|c|c|} 
		\hline
		\bf Dataset & \bf Precision & \bf Recall & \bf F-measure \\  
		\hline
		 User 04 & 0.91 & 0.30 & 0.45 \\ 
		\hline
		User 23 & 0.83 & 0.84 & 0.83 \\  
		\hline
		User 26 & 0.89 & 0.51 & 0.65 \\ 
		\hline
		User 33 & 0.80 & 0.85 & 0.80 \\ 
		\hline
		User 51 & 0.78 & 0.78 & 0.78 \\ 
		\hline
	\end{tabular}
\end{table}

\begin{table}[htbp!]
	\centering
	\caption{The accuracies of our proposed dynamic threshold based approach}
	\label{our-approach}
	\begin{tabular}{|c|c|c|c|} 
		\hline
		\bf Dataset & \bf Precision & \bf Recall & \bf F-measure \\  
		\hline
		User 04 & 0.89 & 0.70 & 0.78 \\ 
		\hline
		User 23 & 0.84 & 0.84 & 0.84 \\ 
		\hline
		User 26 & 0.92 & 0.72 & 0.80 \\ 
		\hline
		User 33 & 0.82 & 0.86 & 0.81 \\ 
		\hline
		User 51 & 0.86 & 0.85 & 0.85 \\ 
		\hline		
	\end{tabular}
\end{table}

\begin{figure}[htbp!]
	\centering
	\includegraphics[height=5cm,keepaspectratio]{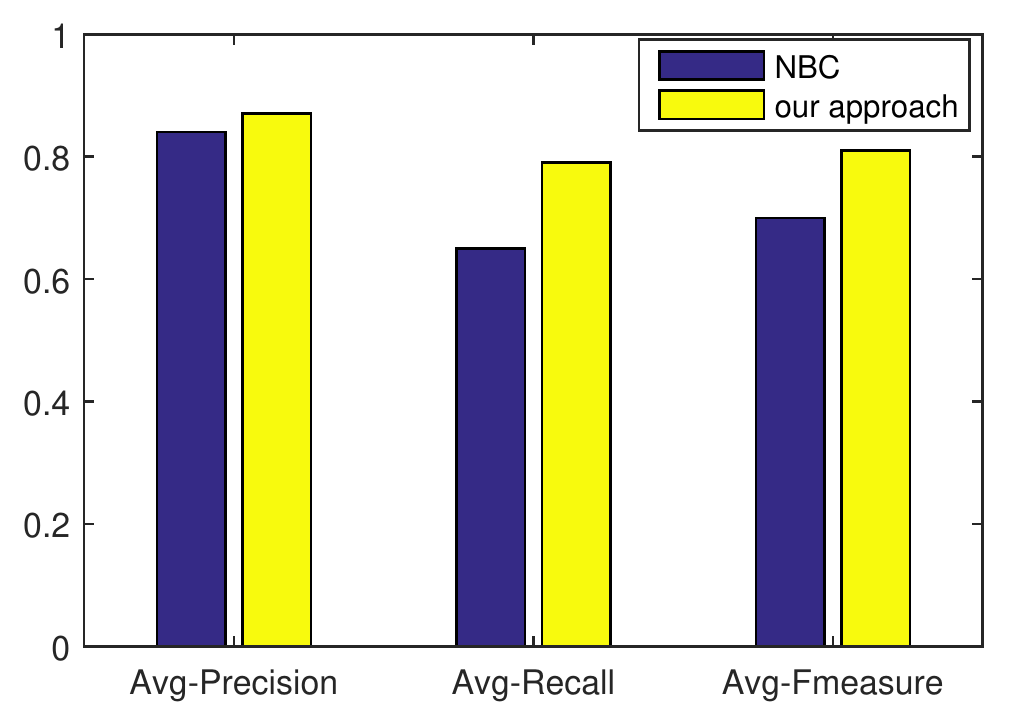}
	\caption{Effectiveness comparison results}
	\label{fig:comparison}
\end{figure}

Table \ref{existing-naive} and Table \ref{our-approach} show the experimental results for five individual mobile phone users' datasets using the existing naive Bayes classifier based noise detection approach and our dynamic threshold based approach respectively. From Table \ref{existing-naive} and Table \ref{our-approach}, we find that our approach consistently outperforms previous NBC-based approach for all individuals in terms of precision, recall and F-measure. In addition to compare individual level, we also show the relative comparison of average precision, average recall and average F-measure for all the five different datasets in Figure \ref{fig:comparison}.

The experimental results for a collection of users show that our approach consistently outperforms the NBC-based approach. The reason is that instead of treating all misclassified instances as noise we identify true noisy instances from misclassified list using a noise threshold. We determine this noise threshold for each individual dataset as it varies according to individual's unique behavioral patterns. As a result, our technique improves the classification accuracy while classifying phone call behavior of individual mobile phone users.

\section{Conclusion and Future Work}
\label{Conclusion and Future Work}
In this paper, we have presented an approach to detecting and eliminating noisy instances from mobile phone data in order to improve the classification accuracy. Our approach dynamically determines the noise threshold according to individual's behavioral patterns. For this, we employ both the naive Bayes classifier and Laplacian estimator. Experimental results on multi-contextual phone call log datasets indicate that compare to the NBC-based approach, our approach improves the classification accuracy in terms of precision, recall and F-measure. 

In future work, we plan to investigate the effect of noise on confidence threshold to produce association rules. We will extend our noise detection technique to produce confidence-based association rules of individual mobile phone users in multi-dimensional contexts.

\bibliographystyle{plain}
\bibliography{bib-file/noise-bibfile}
\nocite{*}

\end{document}